\documentclass[conference]{IEEEtran}
\makeatletter
\def\ps@headings{%
\def\@oddhead{\mbox{}\scriptsize\rightmark \hfil \thepage}%
\def\@evenhead{\scriptsize\thepage \hfil \leftmark\mbox{}}%
\def\@oddfoot{}%
\def\@evenfoot{}}
\makeatother
\usepackage[justification=centering]{caption}
\usepackage{url}
\usepackage{booktabs}
\usepackage{graphicx,subfigure}
\usepackage{tikz}
\usetikzlibrary{shapes,arrows}
\usepackage{epstopdf}
\usepackage{amsmath}
\usepackage{algorithm}
\usepackage{algpseudocode}
\usepackage{amsmath}
\usepackage{amssymb}
\usepackage{amsthm}
\usepackage{epsfig}

\usepackage{amsfonts}
\usepackage{multirow}
\usepackage{color}
\usepackage{array}
\usepackage{listings}
\usepackage{hyperref}
\usepackage[underline=true]{pgf-umlsd}

\usepackage{setspace}

\begin{document}

\title{Augmenting Automation: Intent-Based User Instruction Classification with Machine Learning}

\author{
    \IEEEauthorblockN{Lochan Basyal and Bijay Gaudel}
    \IEEEauthorblockA{\textit{Stevens Institute of Technology, Hoboken, NJ, USA} \\
    \{lbasyal, bgaudel\}@stevens.edu}
}

\maketitle

\begin{abstract}
Electric automation systems offer convenience and efficiency in controlling electrical circuits and devices. Traditionally, these systems rely on predefined commands for control, limiting flexibility and adaptability. In this paper, we propose a novel approach to augment automation by introducing intent-based user instruction classification using machine learning techniques. Our system represents user instructions as intents, allowing for dynamic control of electrical circuits without relying on predefined commands. Through a machine learning model trained on a labeled dataset of user instructions, our system classifies intents from user input, enabling a more intuitive and adaptable control scheme. We present the design and implementation of our intent-based electric automation system, detailing the development of the machine learning model for intent classification. Experimental results demonstrate the effectiveness of our approach in enhancing user experience and expanding the capabilities of electric automation systems. Our work contributes to the advancement of smart technologies by providing a more seamless interaction between users and their environments.

\end{abstract}

\begin{IEEEkeywords}
Intent Classification, Electric Automation, Machine Learning
\end{IEEEkeywords}

\section{Introduction}
Electric automation systems have transformed the way we interact with our environments, offering unprecedented convenience and efficiency in controlling electrical circuits and devices. Traditional approaches to automation often rely on predefined commands \cite{lbasyal} or manual programming, limiting the adaptability and responsiveness of the system to user needs and preferences. In recent years, there has been a growing interest in developing more intuitive and user-friendly automation systems that can understand and interpret natural language instructions.

In this paper, we present a novel approach to augmenting electric automation systems through intent-based user instruction classification using machine learning techniques. Unlike traditional systems that require users to memorize specific commands or sequences, our approach enables users to communicate their intentions naturally, allowing for more dynamic control of electrical circuits. The key innovation of our approach lies in representing user instructions as intents, which encapsulate the underlying purpose or goal of the user's command. By classifying user intents based on their instructions, our proposed system can interpret and execute the appropriate actions without relying on predefined commands. This not only enhances the user experience but also improves the system's adaptability to new commands and user preferences over time.

Central to our approach is the development of a machine learning model trained on a labeled dataset of user instructions. Leveraging natural language processing techniques and supervised learning algorithms, our model learns to recognize patterns and infer the underlying intent behind each instruction. This enables our proposed system to understand a wide range of user instructions and respond accordingly, regardless of variations in syntax or phrasing. By providing a more seamless and natural interaction between users and their environments, our work contributes to the advancement of smart technologies and opens new possibilities for the future of electric automation.

Furthermore, our paper is organized in such a way that it demonstrates the proposed system in Section II; the datasets used for the intent classification task are discussed in Section III; and the machine learning algorithms used are presented in Section IV. The model development and training are discussed in Section V, the evaluation metrics used for the intent classification are discussed in Section VI, and the results and observations in Section VII demonstrate the graphical representation of accuracy over epochs, loss over epochs, classification report, confusion matrix, and model inference with a regularized model. The paper concludes with the future work in Section VIII.

\section{Proposed System}
\begin{figure}[!t]
\centering
\begin{tikzpicture}[node distance=1.5cm, every node/.style={draw, align=center}]
\node (user) [rectangle] {User Instruction Input};
\node (classify) [rectangle, below of=user] {Intent Classification};
\node (match) [rectangle, below of=classify] {Predefined Intent Matching};
\node (program) [rectangle, below of=match] {Embedded System Programming};
\node (execute) [rectangle, below of=program] {Execution of Operations};
\node (feedback) [rectangle, below of=execute] {Feedback Loop};

\draw [->] (user) -- (classify);
\draw [->] (classify) -- (match);
\draw [->] (match) -- (program);
\draw [->] (program) -- (execute);
\draw [->] (execute) -- (feedback);
\end{tikzpicture}
\caption{Proposed System Workflow}
\label{fig:workflow}
\end{figure}
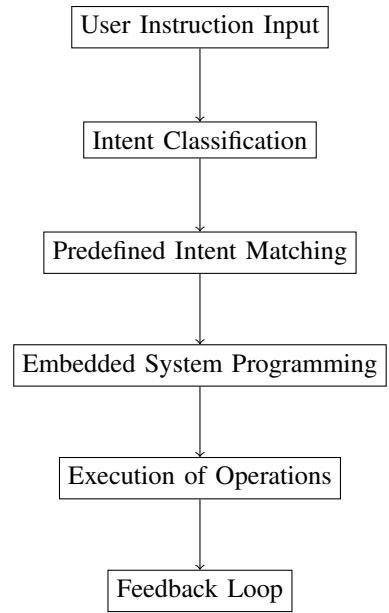

Our proposed system aims to enhance electric automation through the integration of machine learning techniques for intent classification, facilitating more natural and intuitive interaction between users and the automation system. The system workflow can be divided into several key steps, as shown in Figure \ref{fig:workflow}:

\subsection{User Instruction Input}
The system begins with the user providing instructions in the form of text. These instructions can encompass a wide range of commands, expressing the user's intentions regarding the control of electrical circuits, robotic movement, or automation tasks.

\subsection{Intent Classification}
The user-provided instructions are then processed through a machine learning model for intent classification. In our implementation, we utilize Long Short-Term Memory (LSTM) \cite{sepp} networks, a type of recurrent neural network (RNN), for this purpose. Prior to classification, the text instructions undergo preprocessing, including tokenization and TF-IDF (Term Frequency-Inverse Document Frequency) vectorization to convert them into a numerical representation suitable for input into the LSTM model.

\subsection{Predefined Intent Matching}
The classified intents are compared with predefined intents stored within the system. These predefined intents represent a set of commands or actions that the system is capable of executing. The comparison process involves evaluating the equality between the classified intent and the predefined intent to determine the most suitable match.

\subsection{Embedded System Programming}
Upon matching the classified intent with a predefined intent, the corresponding operation is programmed into the embedded controller. The embedded controller serves as the interface between the software-based intent classification system and the physical world of electrical appliances and robotics peripherals. It translates the high-level intent into low-level commands that can be executed by the electronic control system.

\subsection{Execution of Operations}
The electronic control system receives the commands from the embedded controller and executes the corresponding operations in the real world. This may involve turning on/off electrical appliances, adjusting robotic movements, or performing automation tasks as the user's intent dictates.

\subsection{Feedback Loop}
Throughout the process, the system may provide feedback to the user regarding the successful execution of the intended operation or any errors encountered. This feedback loop ensures transparency and user awareness of the system's actions.

The proposed system focuses primarily on the machine learning aspect, specifically intent classification while leveraging existing embedded systems and electronic control mechanisms for physical operation. By employing advanced machine learning techniques, our system enhances the adaptability and user-friendliness of electric automation, paving the way for more intelligent and responsive automation systems.

\section{Datasets}

\begin{figure}[htbp]
    \centering
    \includegraphics[width=1\columnwidth]{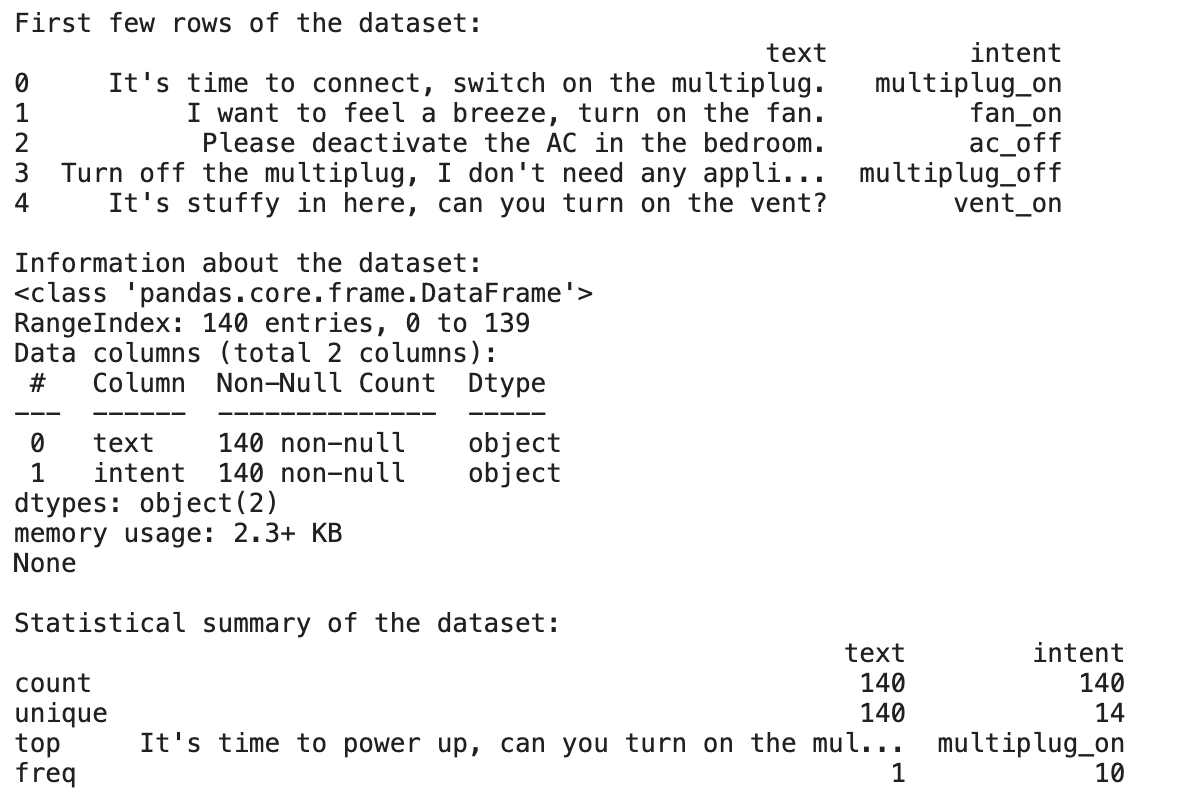}
    \caption{Data Exploration}
    \label{fig:data_exploration}
\end{figure}

For the purpose of training and evaluating our intent classification model for electric automation, we curated a dataset consisting of intent-based user instructions. The dataset comprises a total of 14 intents, each associated with approximately 10 user instructions, resulting in a total of 140 instructions for electric automation. The intents were carefully selected to cover a diverse range of control commands and actions commonly encountered in electric automation scenarios. These intents include commands for turning on/off electrical appliances. Each user instruction in the dataset is labeled with its corresponding intent, allowing the model to learn the mapping between input instructions and their intended actions. Data exploration (refer to Figure \ref{fig:data_exploration}) reveals the first few rows of the dataset, along with information and statistical summaries.

It is important to note that while our dataset provides a foundational representation of intent-based user instructions for electric automation, it is relatively small in scale. As such, the performance of the intent classification model may be limited by the size and diversity of the dataset. To address this limitation and improve the robustness of the model, future research efforts will focus on augmenting the dataset with additional user instructions and intents. By expanding the scope and variety of the training data, we aim to enhance the model's ability to accurately classify a wider range of user commands and accommodate varying user preferences and expressions.

\section{Machine Learning Algorithms}

In our research, we employed a combination of machine learning algorithms to develop an effective intent classification model for electric automation. The key algorithms utilized in our approach include TF-IDF (Term Frequency-Inverse Document Frequency) for text vectorization, one-hot encoding for intent representation, and Long Short-Term Memory (LSTM) networks for sequence modeling.

\subsection{TF-IDF for Text Vectorization}
Prior to training the intent classification model, the user instructions undergo preprocessing, including tokenization and removal of stop words and punctuation. Subsequently, the TF-IDF \cite{tfidf} algorithm is applied to convert the preprocessed text into numerical vectors. TF-IDF assigns weights to terms based on their frequency in a document relative to their frequency across all documents in the dataset. This enables the model to capture the importance of each term in distinguishing between different intents.

\subsection{One-Hot Encoding for Intent Representation}
In parallel with text vectorization, intents are represented using one-hot encoding \cite{onehotencoder}. Each intent is mapped to a binary vector, where a value of 1 indicates the presence of the intent and 0 indicates its absence. This binary representation allows the model to categorically classify each user instruction into one of the predefined intents.

\subsection{Long Short-Term Memory (LSTM) Networks}
To capture the temporal dependencies and semantic context of user instructions, we employed LSTM networks \cite{sepp}, a type of recurrent neural network (RNN). LSTM networks are well-suited for handling sequences of data and are capable of learning long-term dependencies, making them particularly effective for sequence modeling tasks such as intent classification. By processing user instructions as sequences of tokens, LSTM networks can effectively capture the semantic nuances and contextual information necessary for accurate intent classification.

\subsection{Regularization Techniques}
To enhance the generalization and robustness of the intent classification model, we incorporated regularization techniques such as L2 regularization \cite{l2reg} and dropout. L2 regularization penalizes large weights in the model, helping to prevent overfitting and improve the model's ability to generalize to unseen data. Dropout randomly drops a fraction of the connections between neurons during training, reducing the model's reliance on specific features and enhancing its resilience to noise and variability in the input data.

By leveraging these machine learning algorithms in combination, we developed an effective and efficient intent classification model capable of accurately interpreting user instructions for electric automation tasks.

\section{Model Development and Training}
For the development of our baseline intent classification model, we chose to utilize a Long Short-Term Memory (LSTM) network due to its effectiveness in handling long-term sequential data and capturing semantic dependencies within user instructions. The LSTM architecture is well-suited for tasks requiring the retention of context over extended periods, making it an ideal choice for processing natural language inputs. The model \ref{fig:loss_over_epochs} consists of an LSTM layer with 128 units followed by a dense output layer with softmax activation, facilitating multi-class classification. During training, the model is optimized using the categorical cross-entropy loss function and the Adam optimizer. We trained the model for 75 epochs using a batch size of 16 and evaluated its performance on the test dataset.

\begin{figure}[htbp]
    \centering
    \includegraphics[width=0.9\columnwidth]{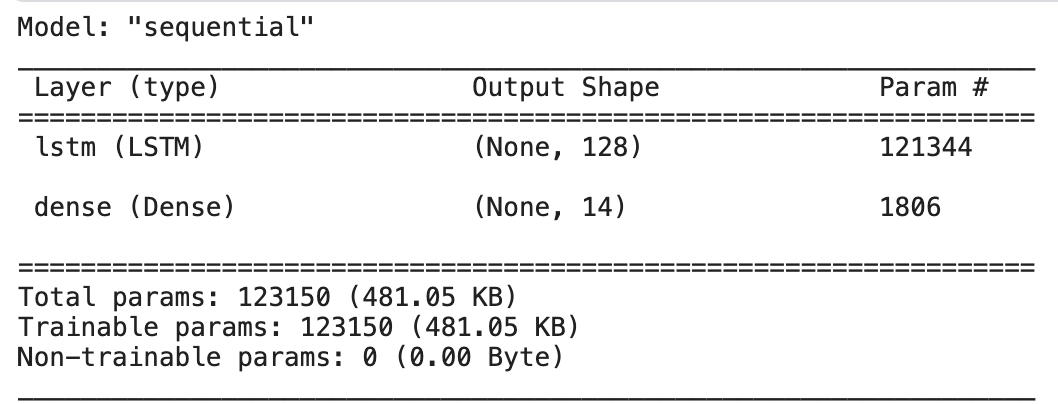}
    \caption{Baseline Model}
    \label{fig:baseline_model}
\end{figure}

\begin{figure}[htbp]
    \centering
    \includegraphics[width=0.9\columnwidth]{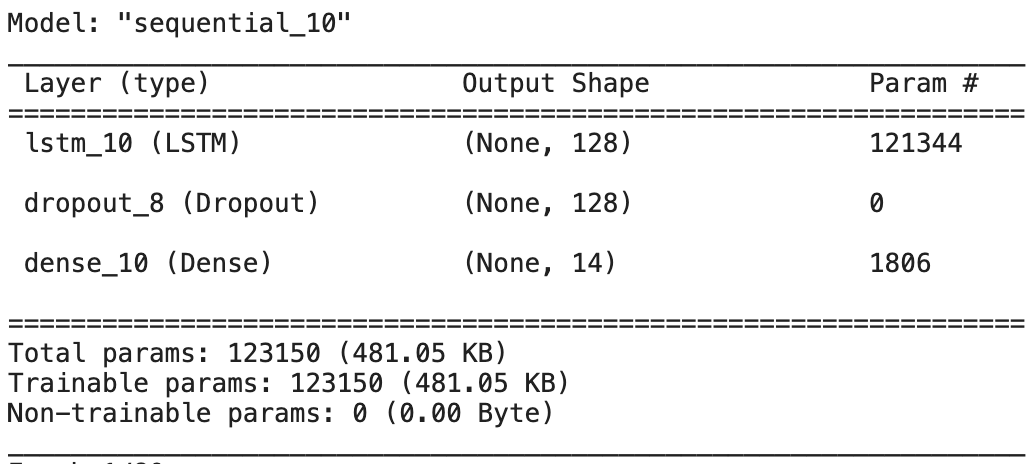}
    \caption{Regularized Model}
    \label{fig:regularized_model}
\end{figure}

In order to enhance the performance and generalization capability of our intent classification model, we incorporated L2 regularization and dropout techniques into the architecture. These regularization techniques help mitigate overfitting and improve the model's ability to generalize to unseen data. In the model development process, L2 regularization with a penalty strength of 0.01 and a dropout rate of 0.2 was applied to prevent overfitting, and the model was trained using the same batch size for 80 epochs. The regularized model is shown in Figure \ref{fig:regularized_model}.

\section{Evaluation Metrics for Intent Classification}

In assessing the performance of our intent classification model for electric automation, we employed a range of evaluation metrics to measure its accuracy and effectiveness in correctly identifying user intents. These evaluation metrics provide insights into the model's overall performance and help assess its suitability for real-world deployment.

\subsection{Accuracy}

Accuracy \cite{prf} measures the proportion of correctly classified user instructions out of the total number of instructions. It provides a general indication of the model's overall correctness in predicting intent and is calculated as: 

\begin{equation}
Accuracy = \frac{{\text{Number of correctly classified instructions}}}{{\text{Total number of instructions}}} \times 100\%
\end{equation}

\subsection{Precision}

Precision \cite{prf} measures the proportion of correctly classified positive predictions (true positives) out of all instances predicted as positive (true positives + false positives). In the context of intent classification, precision reflects the model's ability to accurately identify a specific intent without misclassifying other intents as the target intent. Precision is calculated as:

\begin{equation}
Precision = \frac{{\text{True Positives}}}{{\text{True Positives + False Positives}}}
\end{equation}

\subsection{Recall (Sensitivity)}

Recall \cite{prf} measures the proportion of correctly classified positive instances (true positives) out of all instances that truly belong to the positive class (true positives + false negatives). It indicates the model's ability to capture all instances of a specific intent, without missing any. Recall is calculated as:

\begin{equation}
Recall = \frac{{\text{True Positives}}}{{\text{True Positives + False Negatives}}}
\end{equation}

\subsection{F1 Score}

F1 score \cite{prf} is the harmonic mean of precision and recall, providing a balanced measure of a model's accuracy. It considers both false positives and false negatives and is particularly useful when the class distribution is imbalanced. F1 score is calculated as:

\begin{equation}
F1 = \frac{{2 \times Precision \times Recall}}{{Precision + Recall}}
\end{equation}

\subsection{Confusion Matrix}

The confusion matrix \cite{prf} provides a detailed breakdown of the model's predictions compared to the ground truth labels. It consists of four quadrants: true positives (TP) (correctly classified positive instances), true negatives (TN) (correctly classified negative instances), false positives (FP) (instances that are actually negative but are classified as positive by the model), and false negatives (FN) (instances that are actually positive but are classified as negative by the model). The confusion matrix helps identify specific areas of improvement and provides insights into the model's strengths and weaknesses.

\begin{table}[h]
\centering
\caption{Interpretation of Confusion Matrix}
\label{tab:confusion_matrix_interpretation}
\begin{tabular}{|l|l|l|}
\hline
\textbf{Predicted/Actual} & \textbf{Positive} & \textbf{Negative} \\ \hline
\textbf{Positive}       &  (TP)  &  (FP)       \\ \hline
\textbf{Negative}     &  (FN)   &  (TN)       \\ \hline
\end{tabular}
\end{table}

By evaluating our intent classification model using these metrics, we gain a comprehensive understanding of its performance and can identify areas for refinement and optimization. Additionally, these metrics enable us to compare the effectiveness of different models and approaches, guiding future research and development efforts in electric automation.

\section{Results and Observations}

In this section, we present the results and observations obtained from our experiments on intent classification for electric automation. We compare the performance of two scenarios: a baseline model and a model with regularization and dropout techniques applied. Furthermore, we test the inference of the regularized models on preprocessed test datasets and raw user instructions.

\subsection{Training and Validation Loss Over Epochs}
\begin{figure}[htbp]
    \centering
    \includegraphics[width=0.9\columnwidth]{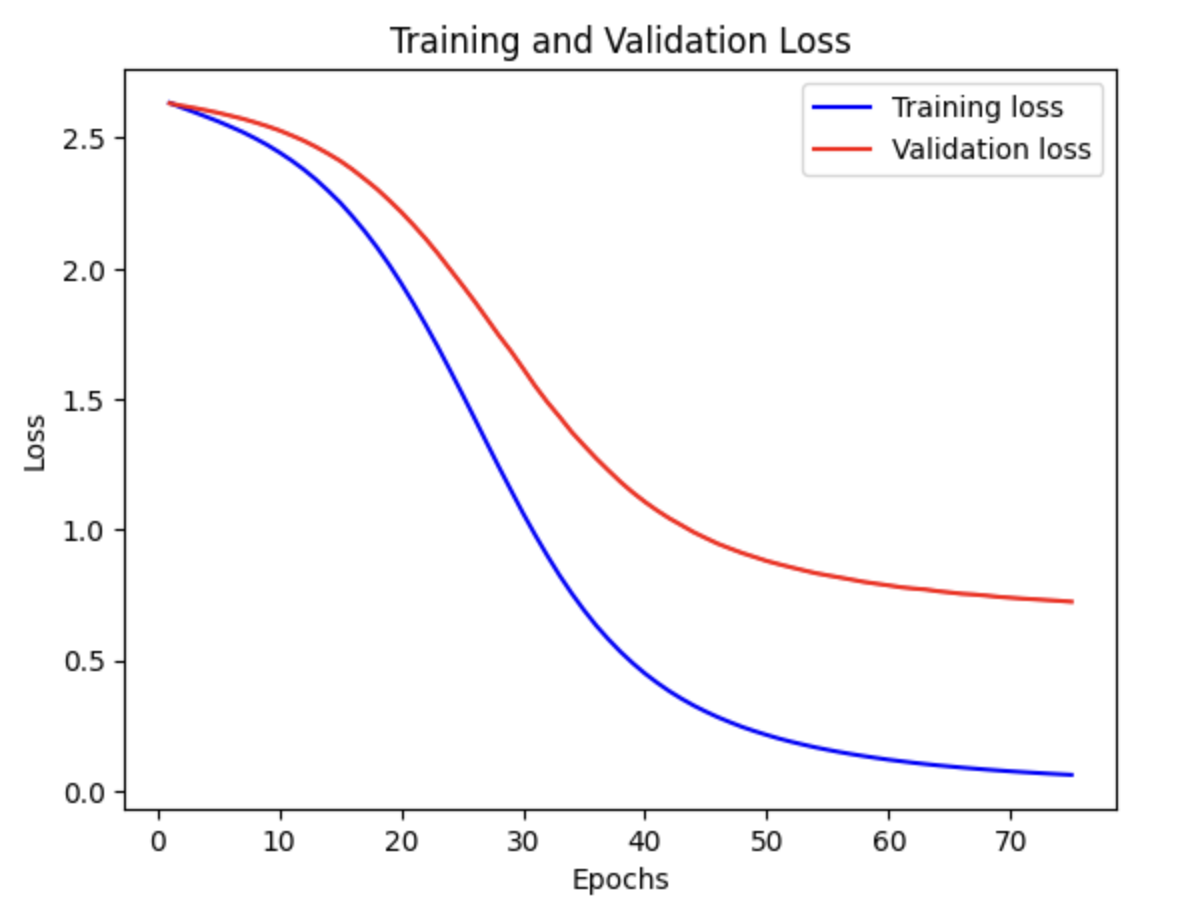}
    \caption{Training and Validation Loss Over Epochs - Baseline Model}
    \label{fig:loss_over_epochs}
\end{figure}

Figure \ref{fig:loss_over_epochs} illustrates the training and validation loss curves for the baseline model, and Figure \ref{fig:loss_over_epochs} represents the regularized model. The validation loss remains consistently higher than the training loss throughout the training process. This widening gap between the training and validation loss suggests that the model may be overfitting to the training data and struggling to generalize well to unseen validation data. However, the model with regularization and dropout exhibits smoother loss curves with reduced fluctuations, suggesting improved stability and generalization.

\begin{figure}[htbp]
    \centering
    \includegraphics[width=0.9\columnwidth]{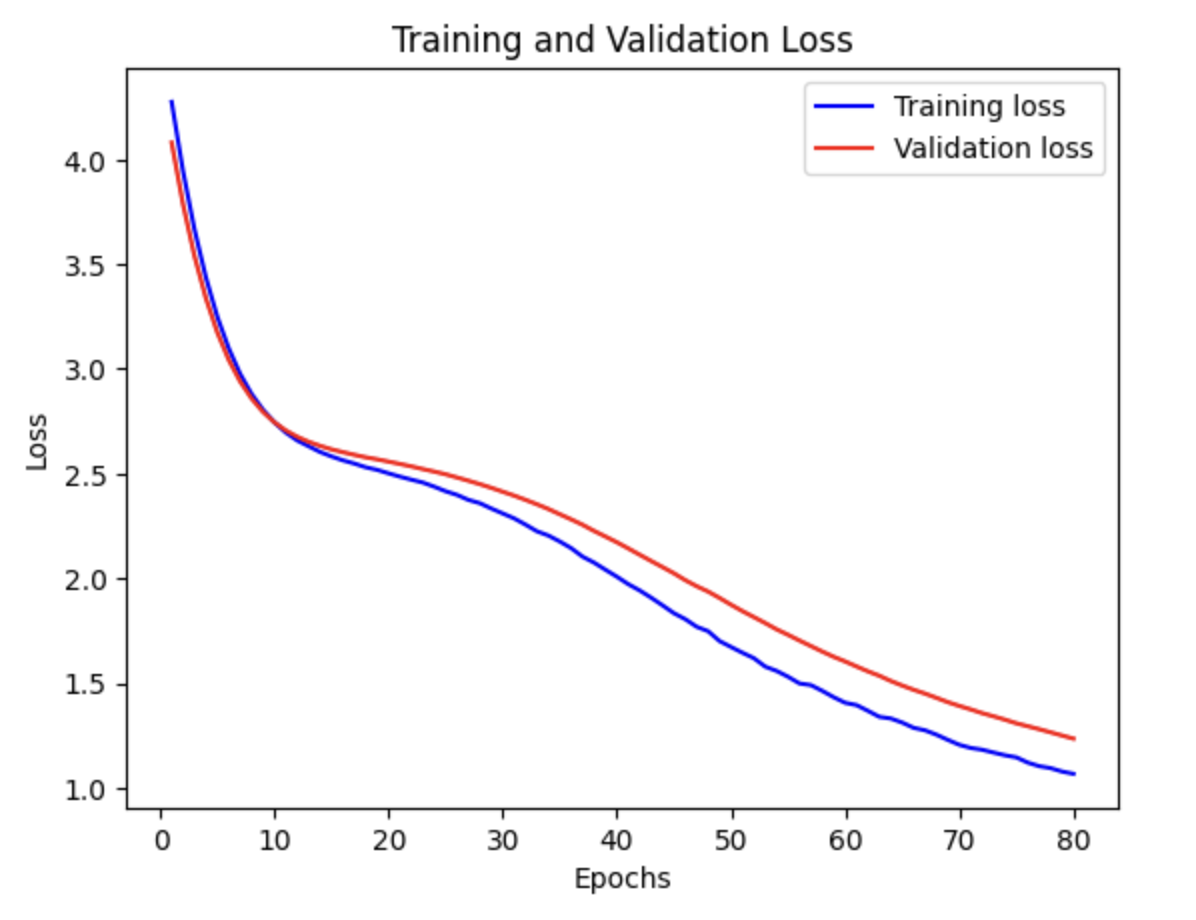}
    \caption{Training and Validation Loss Over Epochs - Regularized Model}
    \label{fig:loss_over_epochs_reg}
\end{figure}

\subsection{Training and Validation Accuracy Over Epochs}
\begin{figure}[htbp]
    \centering
    \includegraphics[width=0.9\columnwidth]{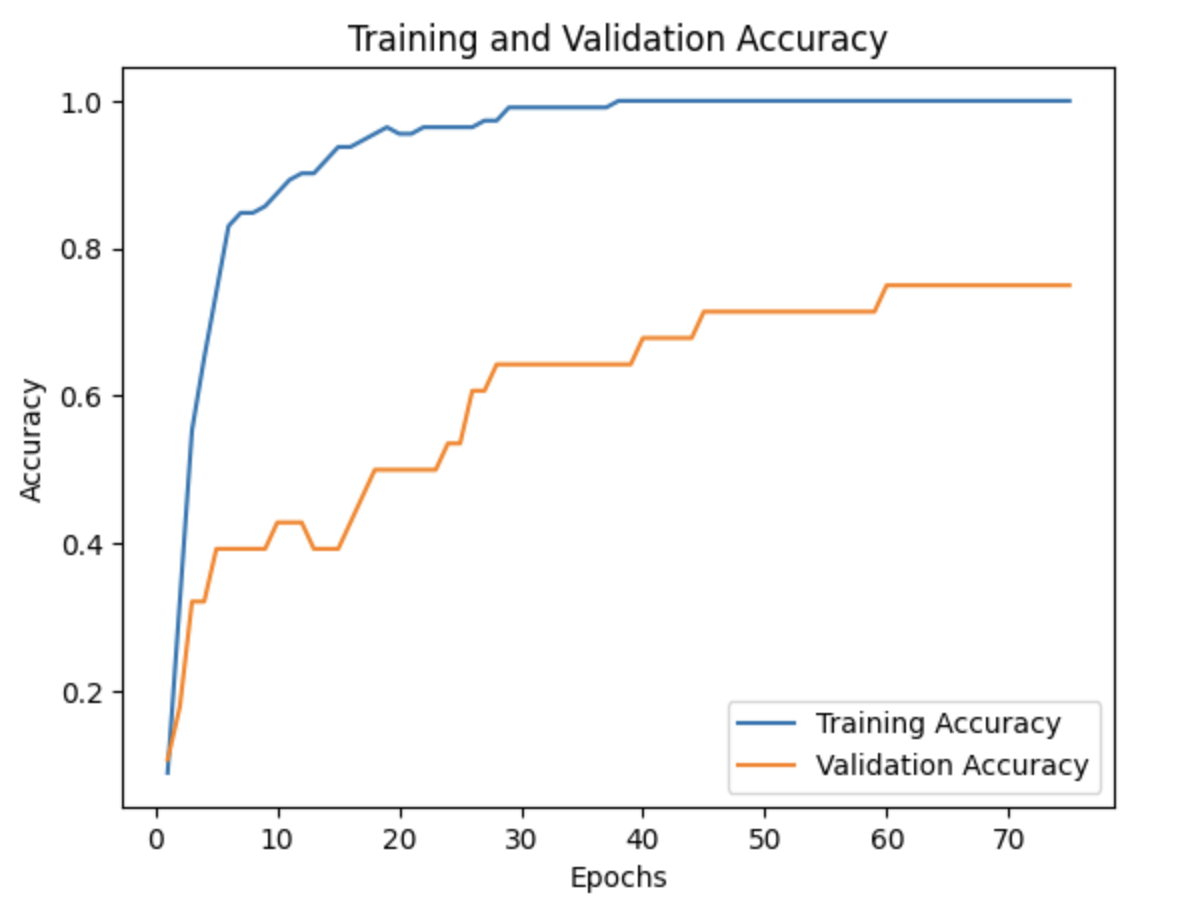}
    \caption{Training and Validation Accuracy Over Epochs - Baseline Model}
    \label{fig:accuracy_over_epochs}
\end{figure}

\begin{figure}[htbp]
    \centering
    \includegraphics[width=0.9\columnwidth]{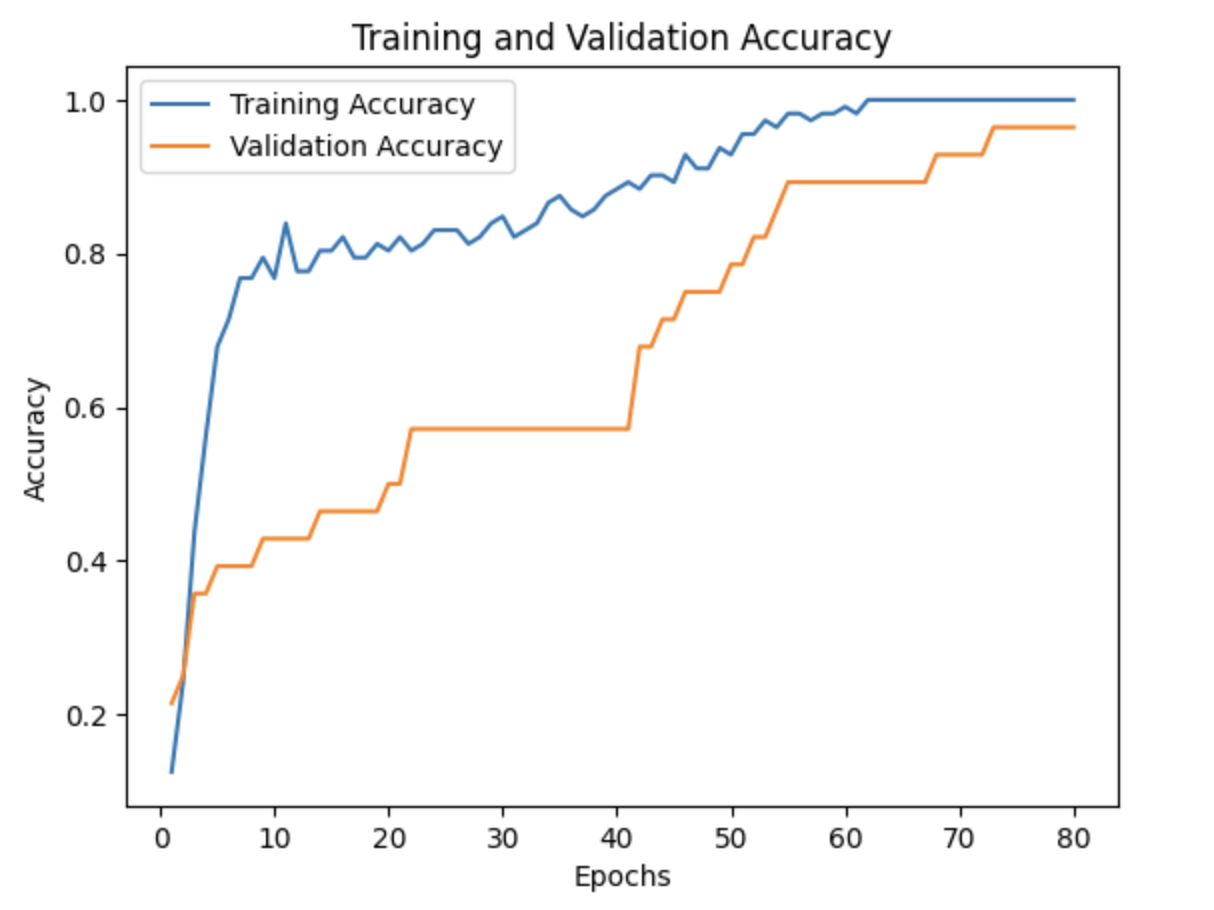}
    \caption{Training and Validation Accuracy Over Epochs - Regularized Model}
    \label{fig:accuracy_over_epochs_reg}
\end{figure}

Figure \ref{fig:accuracy_over_epochs} depicts the training and validation accuracy trends for the baseline model and Figure \ref{fig:accuracy_over_epochs_reg} for the regularized model. The baseline model achieves high training accuracy but shows signs of overfitting, as evidenced by a noticeable gap between training and validation accuracy. In contrast, the model with regularization and dropout achieves comparable training accuracy while maintaining higher validation accuracy, indicating better generalization to unseen data.

\subsection{Classification Report}
\begin{figure}[htbp]
    \centering
    \includegraphics[width=0.9\columnwidth]{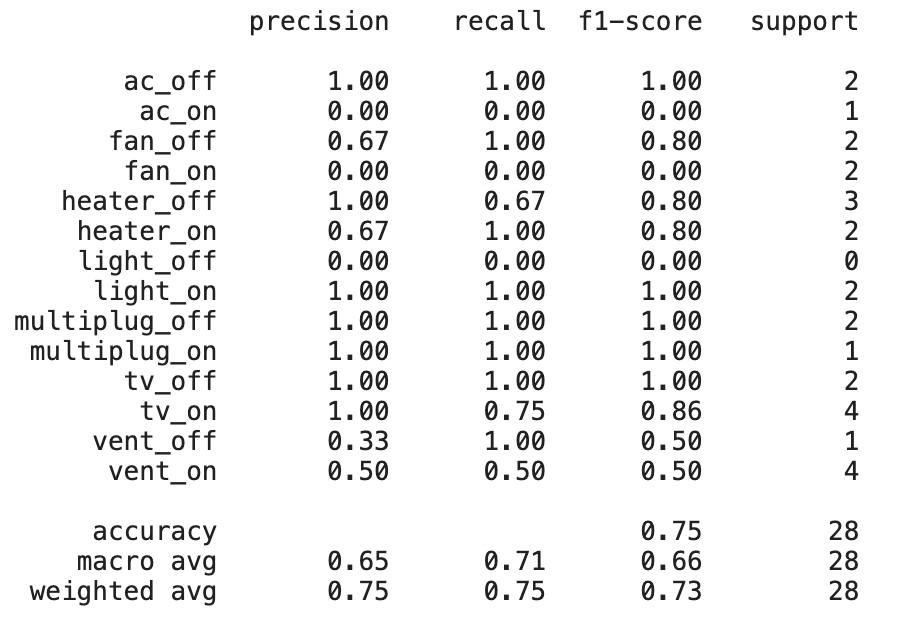}
    \caption{Classification Report - Baseline Model}
    \label{fig:classification_report}
\end{figure}

\begin{figure}[htbp]
    \centering
    \includegraphics[width=0.9\columnwidth]{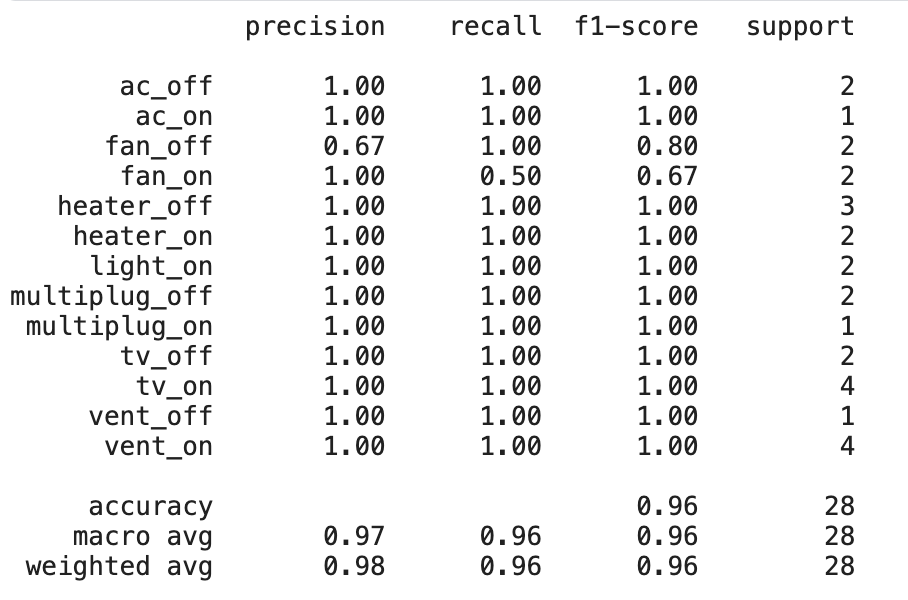}
    \caption{Classification Report - Regularized Model}
    \label{fig:classification_report_reg}
\end{figure}

Figure \ref{fig:classification_report} presents the classification report \cite{creport} for the baseline model, and Figure \ref{fig:classification_report_reg} for the regularized model, including metrics such as precision, recall, and F1-score for each intent class. The model with regularization and dropout demonstrates improvements in precision, recall, and F1-score across most intent classes compared to the baseline model, indicating enhanced performance and robustness. The accuracy obtained with the regularized model is 96\%, compared to the baseline model's 75\%.

\subsection{Confusion Matrix}
\begin{figure}[htbp]
    \centering
    \includegraphics[width=0.9\columnwidth]{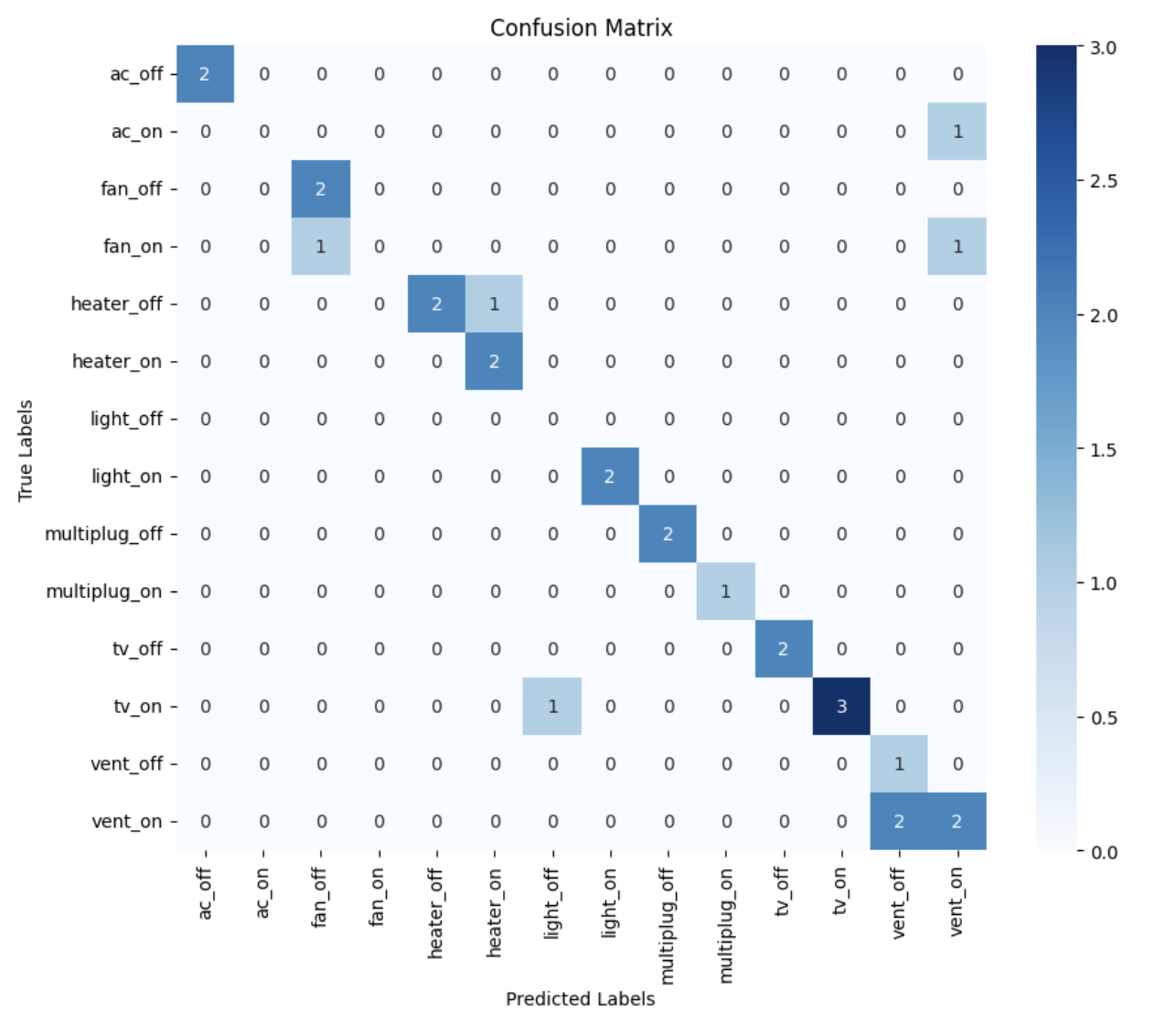}
    \caption{Confusion Matrix - Baseline Model}
    \label{fig:confusion_matrix}
\end{figure}

\begin{figure}[htbp]
    \centering
    \includegraphics[width=0.9\columnwidth]{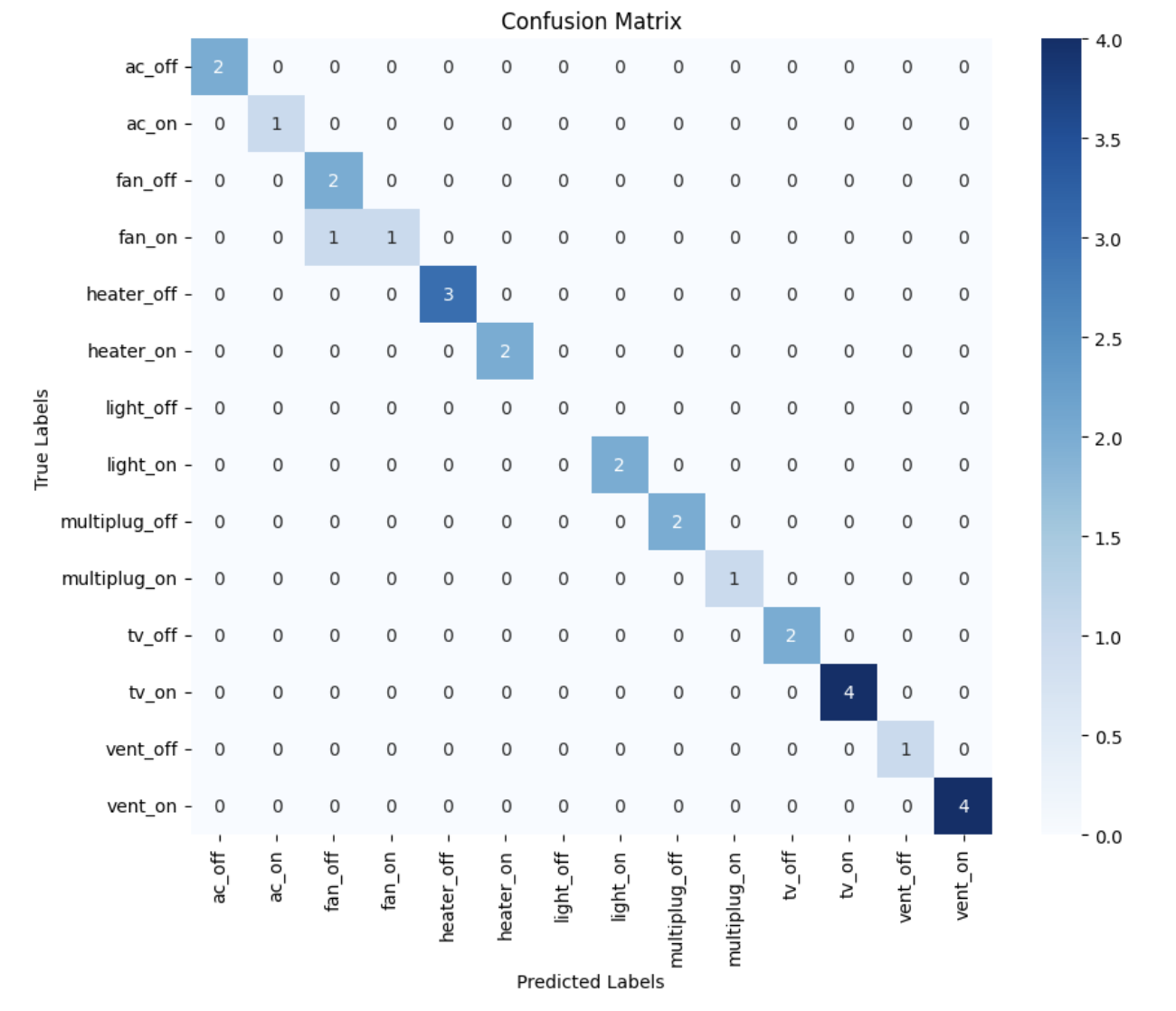}
    \caption{Confusion Matrix - Regularized Model}
    \label{fig:confusion_matrix_regularization}
\end{figure}

The confusion matrices for both scenarios are presented in Figures \ref{fig:confusion_matrix} and \ref{fig:confusion_matrix_regularization}. These matrices provide a detailed breakdown of the model's predictions compared to the ground truth labels, highlighting areas of correct and incorrect classifications for each intent class.

Overall, the experimentation and observation reveal the effectiveness of regularization and dropout techniques in improving the performance and robustness of the intent classification model for electric automation.

\subsection{Model Inference with Regularized Model}
\begin{figure}[htbp]
    \centering
    \includegraphics[width=1\columnwidth]{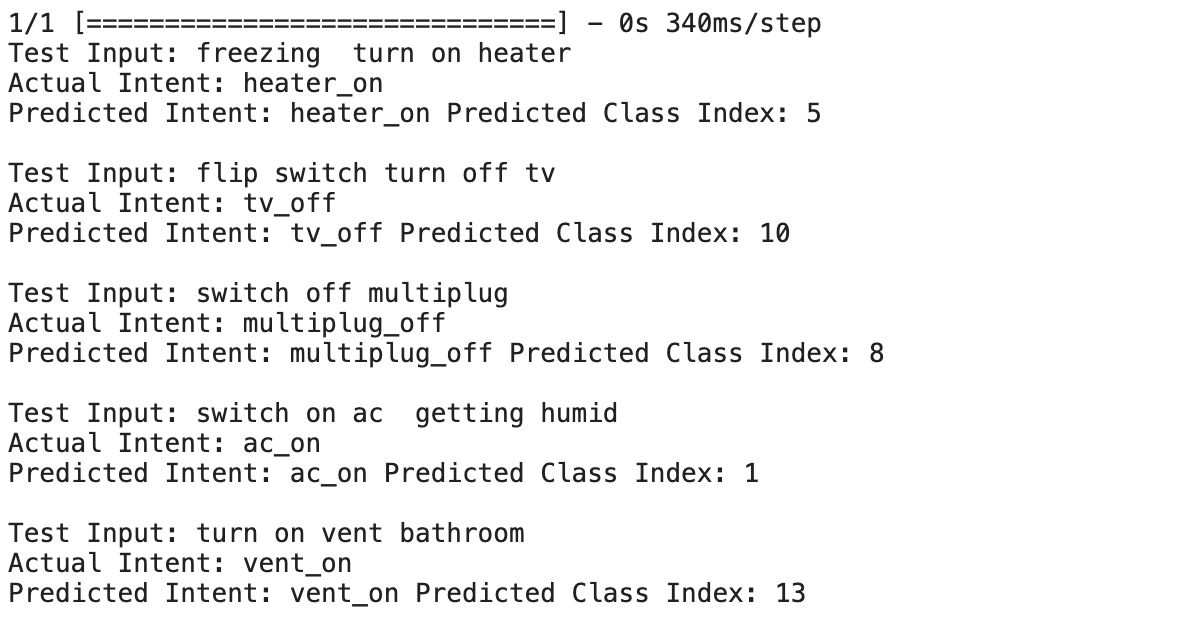}
    \caption{Inference Results for the First 5 Samples of the Preprocessed Test Datasets}
    \label{fig:inference_testing_on_test_datasets}
\end{figure}

We first examine the model's performance on the preprocessed test dataset. The inference results for the first 5 samples are presented in Figure \ref{fig:inference_testing_on_test_datasets}, showcasing the model's ability to classify intents based on the processed user instructions.

\begin{figure}[htbp]
    \centering
    \includegraphics[width=1\columnwidth]{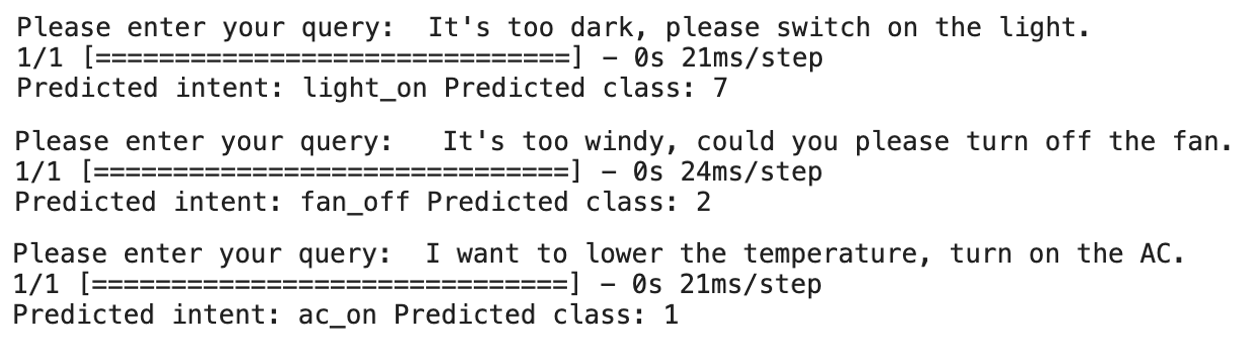}
    \caption{Inference Results from Raw User Instructions}
    \label{fig:inference_testing_on_user_instructions}
\end{figure}

Next, we explore the model's inference capabilities on raw user instructions directly obtained from the users, which is shown in figure \ref{fig:inference_testing_on_user_instructions}. 

\section{Conclusion with Future Work}

In conclusion, the integration of L2 regularization and dropout techniques significantly enhanced the performance of our intent classification model. With the addition of regularization and dropout, the accuracy of the model increased substantially from 75\% to 96\%. This improvement underscores the effectiveness of these techniques in mitigating overfitting and improving the generalization capability of the model.

Moving forward, our research serves as a foundation for the development of an intuitive electrical control system based on intent-based user instructions. Future work will focus on augmenting datasets to further refine the model's performance and adaptability. Additionally, we aim to explore the utilization of language models for intent classification with prompt engineering, enabling more sophisticated interpretation of user instructions.

Furthermore, our efforts will extend towards the development of an end-to-end user-compatible product. This entails integrating communication protocols, embedded systems, and electronic control circuits to create a seamless interface between users and automated systems in the real world. By pursuing these avenues of research, we aim to advance the field of intent-based automation and contribute to the realization of intelligent systems that seamlessly interact with users in various contexts.

\section*{Acknowledgment}
The author would like to acknowledge the foundation laid by his previous work\cite{lbasyal}. This paper represents an iteration aimed at enhancing technology and the user experience with intent-based instructions for electrical automation.

\bibliographystyle{IEEEtran}

\end{document}